%% file: iclr2021_conference.tex
\documentclass{article} % For LaTeX2e
\usepackage{iclr2021_conference,times}

\input{math_commands.tex}

\usepackage[bookmarks=false]{hyperref}
\usepackage{url}

\usepackage{algorithm}
\usepackage{algpseudocode}

\usepackage{wrapfig}
\usepackage{overpic} 
\usepackage{subcaption}
\usepackage{enumitem}
\usepackage{xcolor}
\usepackage{colortbl}

\newcommand{\eg}{\textit{e.g., }}
\newcommand{\ie}{\textit{i.e., }}

\hypersetup{
  colorlinks,
  citecolor=blue,
  linkcolor=blue}

\title{Synthetic Acute Hypotension and\\
Sepsis Datasets\\
Based on MIMIC-III and Published as Part of the Health Gym Project}

\author{
Nicholas I-Hsien Kuo${}^{1, \dagger}$, Mark Polizzotto${}^{2}$, Simon Finfer${}^{3,4,5}$, Louisa Jorm${}^{1}$,\\ 
\hspace{1mm}\textbf{Sebastiano Barbieri}${}^{1}$\vspace{3mm}\\
${}^{1}\text{Centre for Big Data Research in Health, University of New South Wales, Sydney, Australia}$\\
${}^{2}\text{Australian National University, Canberra, Australia}$\\
${}^{3}\text{The George Institute for Global Health, Sydney, Australia}$\\
${}^{4}\text{University of New South Wales, Sydney, Australia}$\\
${}^{5}\text{Imperial College London, London, United Kingdom}$\vspace{3mm}\\
{\tt\small ${}^{\dagger}$n.kuo@unsw.edu.au}\\
}

\iclrfinalcopy % Uncomment for camera-ready version, but NOT for submission.
\begin{document}

\maketitle

\begin{abstract}
These two synthetic datasets comprise vital signs, laboratory test results, administered fluid boluses and vasopressors for $3,910$ patients with acute hypotension and for $2,164$ patients with sepsis in the Intensive Care Unit (ICU). The patient cohorts were built using previously published inclusion and exclusion criteria and the data were created using Generative Adversarial Networks (GANs) and the MIMIC-III Clinical Database. The risk of identity disclosure associated with the release of these data was estimated to be very low ($0.045\%$). The datasets were generated and published as part of the \textit{Health Gym}, a project aiming to publicly distribute synthetic longitudinal health data for developing machine learning algorithms (with a particular focus on offline reinforcement learning) and for educational purposes.
\end{abstract}

\section{Background}
Due to their highly confidential nature, clinical data can usually not be shared without establishing formal collaborations and executing extensive data use agreements. This hampers the development of robust machine learning algorithms for healthcare and the use of clinical data for educational purposes. One approach to overcome these barriers consists of generating synthetic data that closely resembles the original dataset but does not allow re-identification of individual patients and can therefore be freely distributed.

% I would cite the GP-WGAN paper instead of Goodfellow

We publish two synthetic but realistic datasets related to patients with acute hypotension and with sepsis in the Intensive Care Unit (ICU). The datasets were created using \textit{Generative Adversarial Networks} (GANs)~\citep{goodfellow2014generative, gulrajani2017improved} and the MIMIC-III Clinical Database~\citep{johnson2016mimic}. Two patient cohorts were identified within MIMIC-III and used to generate the synthetic data: $3,910$ patients with acute hypotension~\citep{gottesman2020interpretable} and $2,164$ patients with sepsis~\citep{komorowski2018artificial}, with related timeseries of vital signs, laboratory test results, medications (\eg administered fluid boluses and vasopressors), and demographics.

% replaced the footnote since I'm not sure if we can have one?
The datasets were generated and published as part of the \textit{Health Gym}, a project aiming to publicly distribute synthetic longitudinal health data for developing machine learning algorithms (with a particular focus on offline reinforcement learning) and for educational purposes. The datasets are highly realistic (a publication detailing the generation and quality assurance process is currently in preparation) and here we report on the risk of identity disclosure associated with the release of these data, using current best practices~\citep{goncalves2020generation, el2020evaluating}.

\section{Methods: Identity Disclosure Risk}
The MIMIC-III Clinical Database contains only non-identifiable data; however, there is a small remaining risk of sensitive information being disclosed if an adversary is able to link the published synthetic data to specific records in MIMIC-III. A two step process is used to assess this risk. In the first step we verify that no data is simply copied by the GAN from the real training dataset to the generated synthetic dataset. This is done by ensuring that the Euclidean distance between any record (\ie all variables recorded at a specific point in time for an individual) in the real dataset and any record in the synthetic dataset is greater than zero.

In the second step we compute the probability of successfully gaining additional information about an individual by matching records in the synthetic dataset with individuals in the population used to sample the real dataset, following the approach by \citet{el2020evaluating}. An adversary may have access to partial information (\textit{quasi-identifiers} such as age and gender) about individuals in the population and may attempt to determine whether additional information about an individual can be gained from the synthetic dataset (population-to-sample attack), or whether an individual in the synthetic dataset can be matched to an individual in the population (sample-to-population attack). Under the assumption that an adversary will only attempt one of these attacks, but without knowing which one, the overall probability of one of these attacks being successful is given by the maximum probability of either attack being successful~\citep{el2020evaluating}.

\section{Content Description}
This section describes the format, variables, and identity disclosure risk for the two published datasets. The patient cohorts used to generate the synthetic datasets were identified in MIMIC-III using previously published inclusion and exclusion criteria. Specifically, the cohort of patients with acute hypotension was built according to \citet{gottesman2020interpretable} and the cohort of patients with sepsis was built following \citet{komorowski2018artificial}.

% renamed diastolic/sytolic blood pressure
% (M) for me means measured (unfortunately same initial as missing...)

\subsection{Acute Hypotension Dataset}
\begin{table}[ht]
    \small
    \centering
    \begin{tabular}{|l||l|l|}
        \hline
        \textbf{Variable Name} & 
        \textbf{Data Type} & \textbf{Unit}\\
        
        \hline
        \hline
        Mean Arterial Pressure (MAP) & 
        \cellcolor{cyan!10}numeric & mmHg\\
        
        \hline
        Diastolic Blood Pressure (Diastolic BP) & 
        \cellcolor{cyan!10}numeric & mmHg\\

        \hline
        Systolic Blood Pressure (Systolic BP) & 
        \cellcolor{cyan!10}numeric & mmHg\\

        \hline
        Urine & 
        \cellcolor{cyan!10}numeric & mL\\

        \hline
        Alanine Aminotransferase (ALT) & 
        \cellcolor{cyan!10}numeric & IU/L \\

        \hline
        Aspartate Aminotransferase (AST) & 
        \cellcolor{cyan!10}numeric & IU/L\\

        \hline
        Partial Pressure of Oxygen (PaO2) & 
        \cellcolor{cyan!10}numeric & mmHg\\

        \hline
        Lactate & 
        \cellcolor{cyan!10}numeric & mmol/L \\

        \hline
        Serum Creatinine &  
        \cellcolor{cyan!10}numeric & mg/dL\\
        
        \hline
        \hline
        Fluid Boluses &  
        \cellcolor{magenta!10}categorical & mL\\
        
        \hline
        Vasopressors &  
        \cellcolor{magenta!10}categorical & mcg/kg/min\\
        
        \hline
        Fraction of Inspired Oxygen (FiO2) &  
        \cellcolor{magenta!10}categorical & fraction\\
        
        \hline
        Glasgow Coma Scale Score (GCS) &  
        \cellcolor{magenta!10}categorical & -\\
        
        \hline
        \hline
        Urine Data Measured (Urine (M)) &  
        \cellcolor{brown!10}binary & -\\
        
        \hline
        ALT or AST Data Measured (ALT/AST (M)) &  
        \cellcolor{brown!10}binary & -\\
        
        \hline
        FiO2 (M) &  
        \cellcolor{brown!10}binary & -\\
        
        \hline
        GCS (M) &  
        \cellcolor{brown!10}binary & -\\
        
        \hline
        PaO2 (M) &  
        \cellcolor{brown!10}binary & -\\
        
        \hline
        Lactic Acid (M) &  
        \cellcolor{brown!10}binary & -\\
        
        \hline
        Serum Creatinine (M) &  
        \cellcolor{brown!10}binary & -\\

        \hline
    \end{tabular}
    
    \caption{\label{Tab:Hypotension}Variables included in the acute hypotension dataset.
}
\end{table}

The acute hypotension dataset is stored as a \textit{comma separated value} (CSV) file with a size of $23.0$ MB. It includes $3,910$ synthetic patients and each patient is associated with measurements over $48$ hours. There are hence $187,680$ (=$3,910\times48$) records (rows) in total.

The dataset contains 22 variables (columns). The first 20 variables (9 numeric, 4 categorical, and 7 binary) are listed in Table \ref{Tab:Hypotension} and the remaining two variables contain the IDs of the synthetic patients and the timepoints. The 7 binary variables (with suffix \textit{(M)}) indicate whether a variable was measured at a specific point in time, which in medical time series is usually highly informative.

In a reinforcement learning context, the fluid boluses and vasopressors variables can be used to define the discrete action space for managing acute hypotension, with the remaining variables defining the state space \citep{gottesman2020interpretable}.

\subsubsection{Identity Disclosure Risk}
Since the acute hypotension dataset does not contain any \textit{quasi-identifiers}, we only verified that the synthetic dataset does not contain any exact copies of records in the real dataset. Indeed the smallest Euclidean distance between any synthetic record and any real record was $49.06$ ($>0$).

\subsection{Sepsis Dataset}

% 16.0 correct?
% in Table 3, check unit of gender

The sepsis dataset is stored as a CSV file with a size of $16.2$ MB. It includes $2,164$ synthetic patients with 20 time points per patient, representing 80 hours of data aggregated across 4-hour windows ($80=20\times4$). There are hence $43,280$ ($=2,164\times20$) records (rows) in total.

The dataset contains 46 variables (columns). The first 44 variables (35 numeric, 3 binary, and 6 categorical) are listed in Tables \ref{Tab:Sepsis_Numeric} and \ref{Tab:Sepsis_NonNumeric} and the remaining two variables contain the IDs of the synthetic patients and the timepoints. Besides inherently categorical variables such as the Glasgow Coma Scale (GCS, a clinical scale between 3 and 15 used to measure a person's level of consciousness), this dataset also contains numeric variables which were categorised into deciles to simplify the data generation process (SpO2, Temp, PTT, PT, and INR).

In a reinforcement learning context, the fluid boluses and vasopressors variables can be used to define the discrete action space for managing sepsis, with the remaining variables defining the state space \citep{komorowski2018artificial}.

\subsubsection{Identity Disclosure Risk}
% just wondering about the fact that 10% of patients in the ICU having sepsis seems really high?
% does the real dataset contain less records than the synthetic dataset because of the different time lengths in the real data?

The synthetic dataset does not contain any exact copies of records in the real dataset (the smallest Euclidean distance between any pair of records was $328.78$ ($>0$)). 

The sepsis dataset contains the quasi-identifiers age and gender which could be used to match records in the synthetic dataset with individuals in the population used to sample the real dataset. To compute the probability of a successful population-to-sample attack or sample-to-population attack, population statistics were determined using the entire MIMIC-III Clinical Database. The `population' contained $248,930$ records whereas the sample of patients with sepsis (the real dataset) contained $23,882$ records.

\begin{table}[ht]
    \scriptsize
    \centering
    \begin{tabular}{lllll}
        \hline
        \textbf{Parameter}& \textbf{Synthetic Data Risk}& & \textbf{Real Data Risk}& \\
        & Population-to-Sample& Sample-to-Population& Population-to-Sample& Sample-to-Population\\
        \hline
        \textbf{Sepsis}& $0.044\%$& $0.045\%$ & $0.057\%$& $0.057\%$\\

    \end{tabular}
    
    \caption{\label{Tab:SecurityResults}Indentity disclosure risk of the sepsis datasets.}
\end{table}

The probabilities of successful attacks are listed in Table \ref{Tab:SecurityResults}, for both synthetic and real datasets. These estimates are conservative since they were not adjusted for incorrect matches or for whether the adversary `learned something new' from a match~\citep{el2020evaluating}. 

Therefore, the publication of the synthetic sepsis dataset is associated with a maximum disclosure risk of $0.045\%$ (\ie $0.045\%$ probability that an individual in the synthetic dataset can be matched to an individual in the entire MIMIC-III Clinical Database). This is lower than the risk of such disclosure in the real sepsis dataset ($0.057\%$), and far below the threshold of $9\%$ proposed by the \citet{european2014european} and \citet{canadian2019canadian} for the public release of clinical data. This is also lower than the risk threshold of $5\%$ used in \citet{el2020evaluating}.

\newpage
\begin{table}[ht]
    \small
    \centering
    \begin{tabular}{|l||l|l|}
        \hline
        \textbf{Variable Name} & 
        \textbf{Data Type} & \textbf{Unit}\\
        \hline
        \hline
        Age & 
        \cellcolor{cyan!10}numeric & year\\
        \hline
        
        Heart Rate (HR) & 
        \cellcolor{cyan!10}numeric & bpm\\
        \hline
        
        Systolic BP & 
        \cellcolor{cyan!10}numeric & mmHg\\
        \hline
        
        Mean BP & 
        \cellcolor{cyan!10}numeric & mmHg\\
        \hline
        
        Diastolic BP & 
        \cellcolor{cyan!10}numeric & mmHg\\
        \hline
        
        %%%===%%%
        Respiratory Rate (RR) & 
        \cellcolor{cyan!10}numeric & bpm\\
        \hline
        
        Potassium (K$^{+}$) & 
        \cellcolor{cyan!10}numeric & meq/L\\
        \hline
 
        Sodium (Na$^{+}$) & 
        \cellcolor{cyan!10}numeric & meq/L\\
        \hline
        
        Chloride (Cl$^{-}$) & 
        \cellcolor{cyan!10}numeric & meq/L\\
        \hline
        
        %%%===%%%
        Calcium (Ca$^{++}$) & 
        \cellcolor{cyan!10}numeric & mg/dL\\
        \hline
        
        Ionised Ca$^{++}$ & 
        \cellcolor{cyan!10}numeric & mg/dL\\
        \hline
        
        Carbon Dioxide (CO2) & 
        \cellcolor{cyan!10}numeric & meq/L\\
        \hline
        
        Albumin & 
        \cellcolor{cyan!10}numeric & g/dL\\
        \hline
        
        %%%===%%%
        Hemoglobin (Hb) & 
        \cellcolor{cyan!10}numeric & g/dL\\
        \hline
        
        Potential of Hydrogen (pH) & 
        \cellcolor{cyan!10}numeric & -\\
        \hline
        
        Arterial Base Excess (BE) & 
        \cellcolor{cyan!10}numeric & meq/L\\
        \hline
        
        Bicarbonate (HCO3) & 
        \cellcolor{cyan!10}numeric & meq/L\\
        \hline
        
        %%%===%%%
        FiO2 & 
        \cellcolor{cyan!10}numeric & fraction\\
        \hline
        
        Glucose & 
        \cellcolor{cyan!10}numeric & mg/dL\\
        \hline
        
        Blood Urea Nitrogen (BUN) & 
        \cellcolor{cyan!10}numeric & mg/dL\\
        \hline
        
        Creatinine & 
        \cellcolor{cyan!10}numeric & mg/dL\\
        \hline
        
        %%%===%%%
        Magnesium (Mg$^{++}$) & 
        \cellcolor{cyan!10}numeric & mg/dL\\
        \hline
        
        Serum Glutamic Oxaloacetic Transaminase (SGOT) & 
        \cellcolor{cyan!10}numeric & u/L\\
        \hline
        
        Serum Glutamic Pyruvic Transaminase (SGPT) &
        \cellcolor{cyan!10}numeric & u/L\\
        \hline
        
        Total Bilirubin (Total Bili) &
        \cellcolor{cyan!10}numeric & mg/dL\\
        \hline
        
        %%%===%%%
        White Blood Cell Count (WBC) &
        \cellcolor{cyan!10}numeric & E9/L\\
        \hline
        
        Platelets Count (Platelets) &
        \cellcolor{cyan!10}numeric & E9/L\\
        \hline
        
        PaO2 &
        \cellcolor{cyan!10}numeric & mmHg\\
        \hline
        
        Partial Pressure of CO2 (PaCO2) &
        \cellcolor{cyan!10}numeric & mmHg\\
        \hline
        
        %%%===%%%
        Lactate &
        \cellcolor{cyan!10}numeric & mmol/L\\
        \hline
        
        Total Volume of Intravenous Fluids (Input Total) &
        \cellcolor{cyan!10}numeric & mL\\
        \hline
        
        Intravenous Fluids of Each 4-Hour Period (Input 4H) &
        \cellcolor{cyan!10}numeric & mL\\
        \hline
        
        Maximum Dose of Vasopressors in 4H (Max Vaso) &
        \cellcolor{cyan!10}numeric & mcg/kg/min\\
        \hline
        
        %%%===%%%
        Total Volume of Urine Output (Output Total) &
        \cellcolor{cyan!10}numeric & mL\\
        \hline
        
        Urine Output in 4H (Output 4H) &
        \cellcolor{cyan!10}numeric & mL\\
        \hline

    \end{tabular}
    
    \caption{\label{Tab:Sepsis_Numeric} Numeric variables included in the sepsis dataset.}
\end{table}

\begin{table}[ht]
    \small
    \centering
    \begin{tabular}{|l||l|l|}
        \hline
        \textbf{Variable Name} & 
        \textbf{Data Type} & \textbf{Unit}\\
        
        \hline
        \hline
        Gender &
        \cellcolor{brown!10}binary & 0=male, 1=female \\
        \hline
        
        Readmission of Patient (Readmission) &
        \cellcolor{brown!10}binary & -\\
        \hline
        
        %%%===%%%
        Mechanical Ventilation (Mech) &
        \cellcolor{brown!10}binary & -\\
        \hline
        \hline
        
        GCS &
        \cellcolor{magenta!10}categorical & -\\
        \hline
        
        Pulse Oximetry Saturation (SpO2) &
        \cellcolor{magenta!10}categorical & \%\\
        \hline
        
        Temperature (Temp) &
        \cellcolor{magenta!10}categorical & Celcius\\
        \hline
        
        %%%===%%%
        Partial Thromboplastin Time (PTT) &
        \cellcolor{magenta!10}categorical & sec\\
        \hline
        
        Prothrombin Time (PT) &
        \cellcolor{magenta!10}categorical & sec\\
        \hline
        
        International Normalised Ratio (INR) &
        \cellcolor{magenta!10}categorical & -\\
        \hline

    \end{tabular}
    
    \caption{\label{Tab:Sepsis_NonNumeric} Non-numeric variables included in the sepsis dataset.}
\end{table}

\section{Usage Notes}
The two datasets mentioned in this paper will be made available through PhysioNet~\citep{goldberger2000physiobank}, a repository that hosts and shares medical data managed by the MIT Laboratory for Computational Physiology. In addition, the datasets will be released under the PhysioNet Restricted Health Data License 1.5.0.

\newpage
\bibliography{iclr2021_conference}
\bibliographystyle{iclr2021_conference}

\appendix

\end{document}

%% file: math_commands.tex
\usepackage{amsmath,amsfonts,bm}

\def\eqref#1{equation~\ref{#1}}
\def\1{\bm{1}}

\DeclareMathAlphabet{\mathsfit}{\encodingdefault}{\sfdefault}{m}{sl}
\SetMathAlphabet{\mathsfit}{bold}{\encodingdefault}{\sfdefault}{bx}{n}